\documentclass{Interspeech}
\usepackage{adjustbox}
\usepackage{booktabs}

\interspeechcameraready

\title{Length Aware Speech Translation for Video Dubbing}

\author[]{Harveen Singh}{Chadha$^*$}
\author[]{Aswin Shanmugam}{Subramanian$^*$}
\author[]{Vikas}{Joshi}
\author[]{Shubham}{Bansal}
\author[]{Jian}{Xue}
\author[]{Rupeshkumar}{Mehta}
\author[]{Jinyu}{Li}

\affiliation{}{Microsoft}

\email{hchadha@microsoft.com, aswins@microsoft.com}


\usepackage{comment}

\begin{document}

\maketitle
\ifinterspeechfinal
\def\thefootnote{*}\footnotetext{These authors contributed equally to this work}
\fi
\begin{abstract}

In video dubbing, aligning translated audio with the source audio is a significant challenge. Our focus is on achieving this efficiently, tailored for real-time, on-device video dubbing scenarios. We developed a phoneme-based end-to-end length-sensitive speech translation (LSST) model, which generates translations of varying lengths—short, normal, and long—using predefined tags. Additionally, we introduced length-aware beam search (LABS), an efficient approach to generate translations of different lengths in a single decoding pass. This approach maintained comparable BLEU scores compared to a baseline without length awareness while significantly enhancing synchronization quality between source and target audio, achieving a mean opinion score (MOS) gain of 0.34 for Spanish and 0.65 for Korean, respectively.
\end{abstract}

\section{Introduction}
End-to-end (E2E) speech-to-text translation (ST) systems have garnered significant attention due to their advantages over traditional cascaded approaches, which sequentially apply automatic speech recognition (ASR) followed by machine translation (MT) \cite{jia2019leveraging, xue2022large}. By converting speech directly into text in the target language, E2E models reduce the error propagation that is common in cascaded systems, where inaccuracies in ASR can adversely affect the subsequent MT stages \cite{zhang2022revisiting}. This direct approach not only reduces computational cost and latency but also simplifies the system architecture, making deployment and maintenance more efficient compared to managing separate ASR and MT components.


A critical application of speech translation is automatic dubbing, which plays a key role in improving linguistic accessibility in video content. Effective dubbing requires precise synchronization between the translated speech and the original audio in addition to high quality translation. Video dubbing faces challenges due to varying speech durations across languages, complicating temporal alignment. These discrepancies often lead to artificial adjustments that result in unnatural speech. In this work, we focus on the challenge of temporal alignment in video dubbing in the context of E2E ST to ensure that translated speech naturally matches the timing of the source audio. Translated audio length depends on the translated text, created by the ST model, and the text-to-speech (TTS) system's duration model, which controls timing and prosody. 

Several studies have explored length-controlled MT with applications in dubbing \cite{VideoDubber, schioppa2021, karakanta2020, li2022, liu2020, rao2023}. In this work, we introduce a novel yet simplified approach to address alignment challenges in dubbing. Rather than imposing strict duration constraints during translation, we propose a \textbf{Length-Sensitive Speech Translation (LSST)} model, which generates multiple candidate translations of varying lengths—short, normal, and long. Our LSST model outputs an n-best list with varying lengths of translation, providing flexibility in selecting the most suitable translation based on timing constraints. The duration model of the chosen speaker then determines which candidate best aligns with the source audio’s length. 

In the context of automatic dubbing, \cite{MT_verbose_amzn}, \cite{rao2023}, and \cite{tam2022}, have used length tokens to guide translation length for dubbing. While these approaches influence our work, our focus is on real-time, on-device E2E speech translation, where decoding efficiency and latency are critical. To this end, we introduce an optimized decoding strategy called \textbf{Length-Aware Beam Search (LABS)}, which efficiently produces translations of varying lengths within a single decoding pass, which otherwise would require multiple decoding passes to generate translations of varying lengths. 

Another important aspect of our proposed approach is the use of phonemes instead of characters for length modeling. Phoneme-based length ratios provide a more consistent representation across languages compared to character-based methods, making our approach highly scalable and adaptable to diverse linguistic contexts. In \cite{mhaskar2024}, an isometric MT system is introduced, utilizing phoneme count ratio reward-based reinforcement learning to align phonemes between source and target sentences, thereby effectively controlling duration. While their approach also leverages phonemes, it differs from ours in that it relies on reinforcement learning, whereas our LSST model employs predefined length tags for flexible translation control.

Our time-aware speech translation model intelligently adjusts translations by expanding or condensing text as needed, without compromising meaning. By generating translations of varying lengths, our method enhances temporal alignment while avoiding unnatural speed variations, resulting in a more fluid and seamless dubbing experience.


Our key contributions include Length-Sensitive Speech Translation (LSST) with Length-Aware Beam Search (LABS), a translation model that generates multiple candidate outputs of varying lengths (short, normal, long) in an efficient single-pass decoding process.  Existing literature focuses on text-to-text translation, whereas our approach integrates length control, efficient decoding, and phoneme-based length ratios in the realm of end-to-end speech-to-text translation with a focus on real-time on-device dubbing applications, a relatively unexplored area.

\section{Model}
In dubbing, the source and translated audio should align precisely in duration. The duration of translated audio is influenced by: (a) the length of the translated text, and (b) the duration model within the text-to-speech (TTS) system. Prior work \cite{VideoDubber} explored length-controlled translation by conditioning the translation model on absolute and remaining duration. However, with modern human-like conversational TTS systems~\cite{ren2020fastspeech2}, duration is determined based on the entire text, speaker characteristics, and prosodic prompts, making it impossible to know the duration upfront. Additionally, constraining translation to fit a remaining duration often degrades quality by leading to unnatural phrasing.

Our approach addresses alignment by having the speech translation model generate multiple length variants in the n-best list. We then estimate the duration for each translation using a duration model (without generating audio for computational efficiency) and select the variant closest to the source audio length. To ensure alignment does not compromise quality, we apply rules during the final selection process. We next discuss our proposed model which can generate varying length translations.

\subsection{Length Sensitive Speech Translation}
The LSST model is a core component of our approach, designed to generate translations of varying lengths—short, normal, and long—using predefined length control tokens. During training, we prepend one of the length tokens $\{<short>, <normal>, <long>\}$  as a prefix to each target translation, replacing the Start of Sequence (SOS) token. These length tags ($\ell$) are assigned based on the target-to-source length ratio ($r$) in the training data as follows:

\begin{equation}
\ell = \begin{cases} 
<short> & \text{if } r < 1 - \alpha \\ 
<normal> & \text{if } 1 - \alpha \leq r \leq 1 + \alpha \\ 
<long> & \text{if } r > 1 + \alpha.
\end{cases}
\end{equation}


where $\alpha$ is a threshold fixed at $0.1$. We explored computing the length ratio using both character length and phoneme length. Comparing character lengths between languages can be informative, especially when the languages share similar writing systems. For instance, Spanish and English both use the Latin alphabet, making their orthographic representations relatively comparable. However, when comparing languages that use different writing systems, such as English and Korean, the differences become more pronounced. English characters typically correspond to single phonemes, while Korean's Hangul script organizes characters into syllabic blocks, each representing multiple phonemes. This structural distinction means a single Korean character can encapsulate more phonemic information than a single English character, leading to discrepancies when comparing text lengths based solely on character counts. Therefore, phoneme length is more appropriate, and hence we used it.

During inference, the LSST model can generate short, normal, and long variants by conditioning on the respective length token. Unlike conventional speech translation models that produce a single output, LSST provides multiple length-controlled candidates. However, generating all three variants requires separate decoding runs, which is computationally expensive. In the next section, we present an approach to efficiently generate all length variants in a single decoding pass.



\subsection{Length Aware Beam Search (LABS)}\label{sec:figures}

Beam search approximates the highest-probability output sequence $\hat{y}$ by maximizing the conditional probability $P(\hat{y}|x)$ given an input sequence $x$. Standard beam search exhibits two key limitations when applied to LSST models, which are designed to produce translations of varied lengths:

\begin{enumerate}[label=(\arabic*),leftmargin=*]
    \item \textbf{Output Diversity Constraint:} 
    Traditional beam search typically generates n-best lists where candidate sequences differ only marginally. This limited diversity is especially problematic for LSST models, which require a broad exploration of translations across varying lengths.
    
    \item \textbf{Inference Inefficiency:} Obtaining multiple length variations conventionally necessitates running separate inference passes, each conditioned on a distinct length-specific start-of-sequence token (e.g., \([SOS_s]\), \([SOS_n]\), \([SOS_l]\)). This approach can be formulated as:
    
\[
\hat{y}_k = \underset{\hat{y}}{\mathop{\mathrm{argmax}}} \, P(\hat{y} \mid x, \mathrm{SOS}_k) \quad \text{for} \quad k \in \{s, n, l\}
\]
         In this setting, $\hat{y}_k$ denotes the output translation for a specific length category, where $\quad k \in \{s, n, l\}$ corresponds to long, normal, and short translations, respectively. This strategy effectively triples the computational cost during inference.
\end{enumerate}
The challenge, therefore, is to develop a decoding algorithm that efficiently generates a diverse set of translations ($\hat{y}_s$, $\hat{y}_n$, and $\hat{y}_l$) within a single pass through the model.

\subsection{LABS Algorithm}
LABS addresses these issues by modifying standard beam search to explicitly incorporate length tags at the initialization step. Unlike standard beam search, which initializes with a single start token, LABS initializes the beam with the length specific tokens, $[<short>]$, $[<normal>]$, and $[<long>]$, represented as $\mathcal{L} = \{s, n, l\}$, where each $\ell \in \mathcal{L}$ corresponds to a specific desired length.

The algorithm can be formalized as follows:

\begin{enumerate}[label=(\arabic*),leftmargin=*]
    \item \textbf{Initialization:} Instead of a single SOS token, the initial set of hypotheses are initialized as $B_0 = \{ <s> ,<n>, <l>\}$, with associated scores $S_0(\ell) = 0$ for all $\ell \in \mathcal{L}$.

    \item \textbf{Beam Expansion:} At each time step $t$, and for each length tag $\ell \in \mathcal{L}$, new hypotheses are generated by extending each hypothesis $b$ in the length-specific sub-beam $B_t^{(\ell)}$ with each token $v$ from the vocabulary $V$ of the target language. The set of new hypotheses for each length tag is given by:
        \begin{equation}
            B_{t+1}^{(\ell)} = \{ (b \oplus v,   S_{t+1} (\ell))  |  b \in B_t^{(\ell)}, v \in V \}
        \end{equation}
        where $\oplus$ denotes the token append operator, and $S_{t+1}(\ell)$ is the updated score of the extended hypothesis.
        The $S_{t+1}(\ell)$ score is calculated as:
          \begin{equation}
               S_{t+1}(\ell) = S_t(\ell) + \log P(v | b, \ell, x)
           \end{equation}
        The complete beam at time step $t+1$, $B_{t+1}$, is the union of all length-specific sub-beams:
        \begin{equation}
            B_{t+1} = \bigcup_{\ell \in \mathcal{L}} B_{t+1}^{(\ell)}
        \end{equation}


    \item \textbf{Pruning:} Pruning is performed in a length-aware manner to maintain diversity while controlling beam size.  A function \texttt{Prune} is used to select the top-$N$ hypotheses from the combined beam $B_t$ at each time step, ensuring representation from each length category. The pruning process is formalized as:
         \begin{equation}
             B_{t+1} = \texttt{Prune}(B_{t+1}, N, \mathcal{L} )
         \end{equation}
        where $N$ is the beam size and the \texttt{Prune} function operates according to the following principles:
        \begin{enumerate}[label=\arabic*.,leftmargin=1cm]
          \item \textbf{Top-N Selection with Diversity:}  The function ranks all hypotheses in $B_{t+1}$ based on their scores $S_{t+1}(\ell)$ in decreasing order. It then selects the top-$N$ hypotheses while ensuring that at least one hypothesis from each length tag $\ell \in \mathcal{L}$ is included in the selected set, provided such hypotheses exist in $B_{t+1}$. Generally, the top three hypothesis from each length tag are included and the remaining $N$-3 translations are picked based on the score. If there are fewer than $N$ hypotheses in total across all length tags, all available hypotheses are retained.
          \item \textbf{Preservation of Length Diversity:} At least one path originating from each length tag $\ell \in \mathcal{L}$ is preserved in the pruned beam $B_{t+1}$, as long as hypotheses with each tag are still being actively explored.
          \item \textbf{EOS Hypothesis Retention:} Hypotheses that have reached the End-of-Sequence token $\texttt{<EOS>}$ are given priority. If a hypothesis reaches $\texttt{<EOS>}$, it is generally retained in the set of candidate translations and may influence the pruning process to ensure complete translations are considered in the final n-best list.
        \end{enumerate}
    \item \textbf{Iteration:} Steps 2 and 3 are iteratively performed until a maximum length $T$ is reached or all hypotheses in the beam have encountered the $\texttt{<EOS>}$ token, i.e., $t < T$.
   \item \textbf{Final Selection:} The n-best list $\hat{H}$ is selected from the set of all completed hypotheses (those that reached $\texttt{<EOS>}$) such that each length tag is represented, if possible. Let $B_{final}$ be the set of completed hypotheses. The final n-best list is obtained by applying a selection function that prioritizes diversity across length tags:
        \begin{equation}
            \hat{H} = \texttt{SelectNBest}(B_{final}, N, \mathcal{L})
        \end{equation}
        where $\texttt{SelectNBest}(B_{final}, N, \mathcal{L})$ selects up to $N$ hypotheses from $B_{final}$, ensuring representation from each length tag in $\mathcal{L}$ if available in $B_{final}$, and ranking primarily by score.

\end{enumerate}

\section{Experimental Setup}
\subsection{Model and Data}
The ST model used in our experiments is multilingual and jointly trained on Spanish (ES) and Korean (KO) data. The target translation language for the model is English (EN), and the ST model is denoted as (ES, KO) $\rightarrow$ EN.

The S2S model has 24 conformer \cite{gulati2020conformer} encoder blocks and  6 transformer blocks and in total about 120 million parameters. The baseline model without length awareness uses the the same token for both start-of-sequence (SOS) and end-of-sequence (EOS). We use 4332 tokens for English, including $\langle\mathrm{EOS}\rangle$ and $\langle\mathrm{blank}\rangle$. Since we also added the three length tokens for the LSST model, it becomes 4335 tokens in total. 

We trained the model for 16M steps and use a peak learning rate of 2e-4. We used 64 32GB GPUs for training. We use the hybrid CTC-attention loss \cite{watanabe2017hybrid}. We used about 85K hours each of in-house Spanish and Korean audio as the training data for the ST model.  We used FLEURS test set \cite{conneau2022fleurs} for evaluation.  

Subsequent to translation generation with different length tokens from the LSST model, we leverage the duration model learned during the on-device TTS model training as described in the LeanSpeech  architecture \cite{zhang2023leanspeech} to directly estimate the duration of each hypothesis without generating audio. We utilized LeanSpeech for grapheme-to-phoneme (G2P) conversion of the training data, enabling the addition of length tokens for training the LSST model.

To ensure a consistent evaluation paradigm, we processed the source audio by applying a silence removal algorithm, generating a reference duration for each processed audio. This methodology for computing the reference and hypothesis durations for evaluation is consistent with that which is used during real-time deployment and dubbing.


\subsection{Evaluation Metrics}

The Speech Rate Compliance (SRC) metric is defined as the percentage of translations that fall within an acceptable threshold of speech rate variance. We use 20\% as the threshold in our evaluation. The variance is calculated as a ratio between source and translated durations. The translated duration is calculated from the audio synthesized using the TTS systems


We use \texttt{SacreBLEU} \cite{post-2018-call} to compute the bilingual evaluation understudy (BLEU) scores and we use the default \texttt{13a} tokenizer. We also investigate the length ratio (LR) of the predicted translation with respect to the reference translation given by \texttt{SacreBLEU}. 


We conducted a subjective evaluation with four videos each for Spanish and Korean, each video being two minutes long and evaluated by ten judges per video. A mean opinion score (MOS) test was performed, where evaluators were given dubbed videos and asked to assess synchronization quality on a scale of 0 to 5. This test measures the temporal alignment between the audio and visual elements within the system's output.

\section{Experimental Results}
\subsection{Character vs Phoneme as Length Unit in LSST}
\begin{table}[h]
\centering
\begin{adjustbox}{width=1\columnwidth}
\begin{tabular}{c|c|cc|cc}
        \toprule
        \textbf{Method} & \textbf{Length} & \multicolumn{2}{c|}{\textbf{BLEU}} & \multicolumn{2}{c}{\texttt{SacreBLEU} \textbf{LR}} \\
        \cmidrule(r){3-4} \cmidrule(l){5-6}
        & \textbf{Token} & \textbf{Character} & \textbf{Phoneme} & \textbf{Character} & \textbf{Phoneme} \\
        \midrule
\textbf{LSST} & $<short>$  & 22.8 & 22.6 & 1.063 & 1.055 \\ \hline
\textbf{LSST} & $<normal>$ & 22.5 & 22.0 & 1.089 & 1.080 \\ \hline
\textbf{LSST} & $<long>$  & 21.7 & 21.2 & 1.113 & 1.104 \\ \hline
\textbf{Baseline} & N/A & \multicolumn{2}{c|}{22.2} & \multicolumn{2}{c}{1.075}  \\ \hline
\end{tabular}
\end{adjustbox}
\caption{Spanish to English Length Sensitive ST – Length Ratio Criteria Comparison with FLEURS test set}
\label{tab:length_criteria_comparison}
\end{table}


To evaluate the effectiveness of using phonemes as the unit of length for LSST, we conducted experiments on the Spanish (ES) to English (EN) translation task (ES$\rightarrow$EN), allowing us to compare phonemes with characters. The models here only use about 70\% of the available Spanish training data so the results here are not directly comparable to the final model presented in Section \ref{sec:final_results}. The results, presented in Table \ref{tab:length_criteria_comparison}, indicate that both character-based and phoneme-based LSST models achieve similar BLEU scores to the baseline when using the $<normal>$ length token. This demonstrates that LSST models maintain translation quality.

Additionally, the length ratio of the number of tokens in the translated output with respect to the number of tokens in the reference translation is shown. There is a significant increase in the length ratio  from $<short>$ to $<long>$ with the LSST models, indicating the model's effectiveness in controlling output length based on the length token. The trends in BLEU scores and length ratios are consistent for both phoneme-based and character-based LSST models, establishing that phoneme length ratios perform similarly to character length ratios. This finding enabled us to extend the method to languages like Korean

\subsection{Phoneme-based LSST Model with LABS}
\label{sec:final_results}


\begin{table}[h]
\centering
\begin{adjustbox}{width=1\columnwidth}
\begin{tabular}{c|c|cc|cc}
        \toprule
        \textbf{Method} & \textbf{Length} & \multicolumn{2}{c|}{\textbf{SRC}} & \multicolumn{2}{c}{\textbf{BLEU}} \\
        \cmidrule(r){3-4} \cmidrule(l){5-6}
        & \textbf{Token}  & ES & KO & ES & KO \\
        \midrule
                \textbf{Baseline} & N/A &  49.06 & 62.02 & 23.66 & 22.57 \\ \hline
        \textbf{LSST} & $<long>$  & 55.35 & 71.94 & 23.48 & 22.60  \\ \hline
        \textbf{LSST} & $<normal>$ &  52.92 & 67.10 & 23.68 & 23.43   \\ \hline
        \textbf{LSST} & $<short>$  & 48.78 & 57.06 & 24.39 & 23.38   \\ \hline
        \textbf{LABS} & $\mathcal{L}$ & 57.04 & 74.34 & 23.23 & 23.41   \\ \hline
\end{tabular}
\end{adjustbox}
\caption{Speech Rate Compliance (SRC) and BLEU scores for Spanish (ES) and Korean (KO) evaluated with with FLEURS test set. SRC was computed with a 20\% threshold. LABS initializes the beam search with the set of all three length tokens $\mathcal{L}$.}
\label{tab:labs_results}
\end{table}

Table~\ref{tab:labs_results} presents the evaluation results of the multilingual (ES, KO) $\rightarrow$ EN translation model in terms of both SRC and BLEU metrics on the FLEURS test sets. LABS demonstrates superior performance compared to the baseline in terms of synchronization, notably by better matching the source audio's speaking rate. Specifically, LABS achieves a 16.3\% relative improvement in Spanish and a 19.9\% relative improvement in Korean over the baseline in terms of SRC. The results indicate that LABS produces concise and natural translations while improving alignment between the source and target audio. Additionally, the results of the individual length tokens are shown, and we can see that LABS outperforms all three individual length tokens in terms of SRC.

Moreover, LABS outperforms the baseline in terms of translation quality (BLEU) for Korean. Although the BLEU scores for Spanish exhibit a marginal decline, this trade-off is acceptable given the substantial gain in SRC.

To quantify the practical efficiency of the LABS method, a latency test was conducted. Our tests showed that LABS increased the latency on average by only about 4.3\% compared to the traditional beam search. Thus, LABS maintains latency comparable to that of a traditional beam search initialized with a single length token, even while producing three translations in a single pass.

\subsection{Translation Examples with Phoneme-based LSST Model and LABS}
\begin{itemize}

\item Example 1 - the Baseline translation for the sentence "Every player can only speak once and for all." resulted in a speed ratio of 1.3, whereas the LABS-generated translation, "Each player can only speak once," reduced redundancy and achieved a lower speed ratio of 0.93.

\item Example 2 - the Baseline approach produced "Financing is based on tithe, although it accepts financial, personal, and business support," with a speed ratio of 0.79. The LABS-generated translation, "The financing is based on tithes, although they accept financial support, personal support, and business support," enhanced clarity and maintained better structural balance while achieving a speed ratio of 0.95.
\end{itemize}
These examples demonstrate that LABS effectively controls translation length, improves fluency, and enhances speed ratio adherence while maintaining translation quality.

\subsection{MOS Test}
Our subjective evaluation results showed a MOS gain of 0.34 for Spanish and 0.65 for Korean in terms of synchronization quality compared to the baseline without length awareness.

\section{Conclusion}
In this paper, we introduced a phoneme-based end-to-end length-sensitive speech translation (LSST) model tailored for real-time, on-device video dubbing scenarios. Our approach leverages predefined length control tokens to generate translations of varying lengths—short, normal, and long—while maintaining high translation quality. We also proposed the length-aware beam search (LABS) algorithm, which efficiently produces multiple length variants in a single decoding pass, significantly enhancing synchronization quality between source and target audio, while having a minimal impact on latency. Our experimental results demonstrated that the LSST model, combined with LABS, achieves comparable BLEU scores to the baseline while improving speech rate compliance and synchronization quality. This work advances the state-of-the-art in video dubbing by providing a practical and efficient solution for achieving temporal alignment in translated audio, making it highly suitable for real-time applications.

\newpage
\bibliographystyle{IEEEtran}
\bibliography{mybib}

\end{document}